\documentclass[journal]{IEEEtran}

\usepackage{url}
\usepackage{amsfonts}
\usepackage{amsmath,amssymb,graphicx,epsfig}
\usepackage{enumerate,amsthm,epstopdf}
\usepackage{bbm}
\usepackage{dsfont}
\usepackage{algpseudocode}
\usepackage{algorithmicx}
\usepackage{algorithm}
\usepackage{listings}
\usepackage{tikz}
\usetikzlibrary{fit,positioning}
\usepackage{pstricks}
\usepackage{auto-pst-pdf}
\usepackage{pst-node}
\usepackage{comment}
\usepackage{color}

\usepackage{verbatim}
\newcommand{\argmin}{\operatornamewithlimits{argmin}}

\def\bR{{\mathbb R}}

\def\bN{{\mathbb N}}

\def\f0{{\mathbf 0}}

\def\Tr{{\operatorname{Tr}}}

\theoremstyle{definition}

\usepackage[noadjust]{cite}

\ifCLASSINFOpdf
\else
\fi
\begin{document}
%
\title{Online Matrix Factorization via Broyden Updates}
%
%
%

\author{\"Omer Deniz Aky{\i}ld{\i}z
\thanks{The author is with Bogazici University. Email: deniz.akyildiz@boun.edu.tr}}
%
%

\markboth{}%
{Shell \MakeLowercase{\textit{et al.}}: Bare Demo of IEEEtran.cls for Journals}
%



\maketitle

\begin{abstract}
In this paper, we propose an online algorithm to compute matrix factorizations. Proposed algorithm updates the dictionary matrix and associated coefficients using a single observation at each time. The algorithm performs low-rank updates to dictionary matrix. We derive the algorithm by defining a simple objective function to minimize whenever an observation is arrived. We extend the algorithm further for handling missing data. We also provide a mini-batch extension which enables to compute the matrix factorization on big datasets. We demonstrate the efficiency of our algorithm on a real dataset and give comparisons with well-known algorithms such as stochastic gradient matrix factorization and nonnegative matrix factorization (NMF).
\end{abstract}

\begin{IEEEkeywords}
Matrix factorizations, Online algorithms.
\end{IEEEkeywords}

%
\IEEEpeerreviewmaketitle

\section{Introduction}
\IEEEPARstart{P}{roblem} of online factorization of data matrices is of special interest in many domains of signal processing and machine learning. The interest comes from either applications with streaming data or from domains where data matrices are too wide to use batch algorithms. Analysis of such datasets is needed in many popular application domains in signal processing where batch matrix factorizations is successfully applied \cite{LeeSeungNMF}. Some of these applications include processing and restoration of images  \cite{cemgil09-nmf}, source separation or denoising of musical \cite{FevotteISNMF,smaragdis2003nmftranscription} and speech signals \cite{wilson2008speech}, and predicting user behaviour from the user ratings (collaborative filtering) \cite{korenMFrecommender}. Nowadays, since most applications in these domains require handling streams or large databases, there is a need for online factorization algorithms which updates factors only using a subset of observations.

Formally, matrix factorization is the problem of factorizing a data matrix $Y\in \bR^{m\times n}$ into \cite{LeeSeungNMF},
\begin{align}\label{NMFproblem}
Y \approx C X
\end{align}
where $C \in \bR^{m\times r}$ and $X \in \bR^{r\times n}$. Intuitively, $r$ is the approximation rank which is typically selected by hand. These methods can be interpreted as dictionary learning where columns of $C$ defines the elements of the dictionary, and columns of $X$ can be thought as associated coefficients. \textit{Online} matrix factorization problem consists of updating $C$ and associated columns of $X$ by only observing a subset of columns of $Y$ which is the problem we are interested in this work.

In recent years, many algorithms were proposed to tackle online factorization problem. In \cite{YildirimSMCNMF}, authors propose an algorithm which couples the expectation-maximization with sequential Monte Carlo methods for a specific Poisson nonnegative matrix factorization (NMF) model to develop an online factorization algorithm. The model makes Markovian assumptions on the columns of $X$, and it is similar to the classical probabilistic interpretation of NMF \cite{cedric-eusipco2009}, but a dynamic one. They demonstrate the algorithm on synthetic datasets. In \cite{onlineisnmf}, authors propose an online algorithm to solve the Itakura-Saito NMF problem where only one column of data matrix is used in each update. They also provide a mini-batch extension to apply it in a more efficient manner and demonstrate audio applications. In \cite{mairal2010online}, authors propose several algorithms for online matrix factorization using sparsity priors. In \cite{BucakGunselNMF}, authors propose an incremental nonnegative matrix factorisation algorithm based on an incremental approximation of the overall cost function with video processing applications. In \cite{sismanisSGD}, authors implement a stochastic gradient algorithm for matrix factorization which can be used in many different settings.

In this paper, we propose an online algorithm to compute matrix factorizations, namely the online matrix factorization via Broyden updates (OMF-B). We do not impose nonnegativity conditions although they can be imposed in several ways. At each time, we assume only observing a column of the data matrix (or a mini-batch), and perform low-rank updates to dictionary matrix $C$. We do not assume any structure between columns of $X$, and hence $Y$, but it is possible to extend our algorithm to include a temporal structure. OMF-B is very straightforward to implement and has a single parameter to tune aside from the approximation rank.

The rest of the paper is organised as follows. In Section \ref{SecObjFunc}, we introduce our cost function and the motivation behind it explicitly. In Section \ref{SecOnlineAlg}, we derive our algorithm and give update rules for factors. In Section \ref{SecModifications}, we provide two modifications to implement mini-batch extension and update rules for handling missing data. In Section \ref{SecExperiment}, we compare our algorithm with stochastic gradient matrix factorization and NMF on a real dataset. Section \ref{SecConc} concludes this paper.
\section{The Objective Function}\label{SecObjFunc}
We would like to solve the approximation problem \eqref{NMFproblem} by using only columns of $Y$ at each iteration. For notational convenience, we denote the $k$'th column of $Y$ with $y_{k} \in \bR^m$. In the same manner, we denote the $k$'th column of $X$ as $x_{k} \in \bR^r$ where $r$ is the approximation rank. This notation is especially suitable for matrix factorisations when columns of the data matrix represent different instances (e.g. images). We set $[n] = \{1,\ldots,n\}$ for $n\in \bN$.

We assume that we observe \textit{random} columns of $Y$. To develop an appropriate notation, we use $y_{k_t}$ to denote the data vector observed at time $t$ where $k_t$ is sampled from $[n]$ uniformly random. The use of this notation implies that, at time $t$, we can sample any of the columns of $Y$ denoted by $y_{k_t}$. This randomization is not required and one can use sequential observations as well by putting simply $k_t = t$. We denote the estimates of the dictionary matrix $C$ at time $t$ as $C_t$. As stated before, we would like to update dictionary matrix $C$ and a column of the $X$ matrix $x_{k_t}$ after observing a single column $y_{k_t}$ of the dataset $Y$. For this purpose, we make the following crucial observations:
\begin{itemize}
\item[(i)] We need to ensure $y_{k_t} \approx C_t x_{k_t}$ at time $t$ for $k_t\in [n]$,
\item[(ii)] We need to penalize $C_t$ estimates in such a way that it should be ``common to all observations", rather than being overfitted to each observation.
\end{itemize} 
As a result we need to design a cost function that satisfies conditions (i) and (ii) simultaneously. Therefore, for fixed $t$, we define the following objective function which consists of two terms. Suppose we are given $y_{k_t}$ for $k_t\in [n]$ and $C_{t-1}$, then we solve the following optimization problem for each $t$,
\begin{align}\label{Cost}
(x_{k_t}^*, C_t^*) = \argmin_{x_{k_t},C_t} \big\| y_{k_t} - C_t x_{k_t} \big\|_2^2 + \lambda \big\| C_t - C_{t-1} \big\|_{F}^2
\end{align}
where $\lambda \in \bR$ is a parameter which simply chooses how much emphasis should be put on specific terms in the cost function. Note that, Eq.~\eqref{Cost} has an analytical solution both in $x_{k_t}$ and $C_t$ separately.  The first term ensures the condition (i), that is, $y_{k_t} \approx C_t x_{k_t}$. The second term ensures the condition (ii) which keeps the estimate of dictionary matrix $C$ ``common" to all observations. Intuitively, the second term penalizes the change of entries of $C_t$ matrices. In other words, we want to restrict $C_t$ in such a way that it is still close to $C_{t-1}$ after observing $y_{k_t}$ but also the error of the approximation $y_{k_t} \approx C_t x_{k_t}$ is small enough. One can use a weighted Frobenius norm to define a correlated prior structure  on $C_t$ \cite{hennig2013quasi}, but this is left as a future work.
\section{Online Factorization Algorithm}\label{SecOnlineAlg}
For each $t$, we solve \eqref{Cost} by fixing $x_{k_t}$ and $C_t$. In other words, we will perform an alternating optimisation scheme at each step. In the following subsections, we derive the update rules explicitly.
\subsection{Derivation of the update rule for $x_{k_t}$}
To derive an update for $x_{k_t}$, $C_t$ is assumed to be fixed. To solve for $x_{k_t}$, let $G_{k_t}$ denote the cost function such that,
\begin{align*}
G_{k_t} =\big\| y_{k_t} - C_t x_{k_t} \big\|_F^2 + \lambda \big\| C_t - C_{t-1} \big\|_F^2,
\end{align*}
and set $\nabla_{x_{k_t}} G_{k_t} = 0$. We are only interested in the first term. As a result, solving for $x_{k_{t}}$ becomes a least squares problem, the solution is the following pseudoinverse operation [12],
\begin{align}\label{updateX}
x_{k_t} = (C_t^\top C_t)^{-1} C_t^\top y_{k_t},
\end{align}
for fixed $C_t$.
\subsection{Derivation of the update rule for $C_t$}
If we assume $x_{k_t}$ is fixed, the update with respect to $C_t$ can be derived by setting $\nabla_{C_t} G_{k_t} = 0$. We leave the derivation to the appendix and give the update as,
\begin{align}\label{beforeShermanMorrison}
C_t &= (\lambda C_{t-1} + y_{k_t} x_{k_t}^\top) (\lambda I + x_{k_t} x_{k_t}^\top)^{-1},
\end{align}
and by using Sherman-Morrison formula \cite{matrixcookbook} for the term $(\lambda I + x_{k_t} x_{k_t}^\top)^{-1}$, Eq.~\eqref{beforeShermanMorrison} can be written more explicitly as,
\begin{align}\label{updateC}
C_t &= C_{t-1} + \frac{(y_{k_t} - C_{t-1} x_{k_t}) x_{k_t}^\top}{\lambda + x_{k_t}^\top x_{k_t}},
\end{align}
which is same as the Broyden's rule of quasi-Newton methods as $\lambda \to 0$ \cite{hennig2013quasi}. We need to do some subiterations between updates \eqref{updateX}  and \eqref{updateC} for each $t$. As it turns out, empirically, even $2$ inner iterations are enough to obtain a reliable overall approximation error.
\begin{algorithm}[t]
\begin{algorithmic}[1]
\caption{OMF-B}\label{OMFB}
\State Initialise $C_0$ randomly and set $t = 1$.
\Repeat
\State Pick $k_t \in [n]$ at random.
\State Read $y_{k_t} \in \bR^m$
\For{$\text{Iter} = 1:2$}
\begin{align*}
x_{k_t} &= (C_t^\top C_t)^{-1} C_t^\top y_{k_t} \\
C_t &= C_{t-1} + \frac{(y_{k_t} - C_{t-1}x_{k_t})x_{k_t}^\top}{\lambda + x_{k_t}^\top x_{k_t}}
\end{align*}
\EndFor
\State $t \leftarrow t+1$
\Until{convergence}
\end{algorithmic}
\end{algorithm}
\section{Some Modifications}\label{SecModifications}
In this section, we provide two modifications of the Algorithm~\ref{OMFB}. The first modification is an extension to a mini-batch setting and requires no further derivation. The second modification provides the rules for handling missing data.
\subsection{Mini-Batch Setting}
In this subsection, we describe an extension of the Algorithm~\ref{OMFB} to the mini-batch setting. If $n$ is too large (e.g. hundreds of millions), it is crucial to use subsets of the datasets. We use a similar notation, where instead of $k_t$, now we use an index set $v_t \subset [n]$. We denote a mini-batch dataset at time $t$ with $y_{v_t}$. Hence $y_{v_t} \in \bR^{m \times |v_t|}$ where $|v_t|$ is the cardinality of the index set $v_t$. In the same manner, $x_{v_t} \in \bR^{|v_t|\times n}$ denotes the corresponding columns of the $X$. We can not use the update rule \eqref{updateC} immediately by replacing $y_{k_t}$ with $y_{v_t}$ (and $x_{k_t}$ with $x_{v_t}$) because now we can not use the Sherman-Morrison formula for \eqref{beforeShermanMorrison}. Instead we have to use Woodbury matrix identity \cite{matrixcookbook}. However, we just give the general version \eqref{beforeShermanMorrison} and leave the use of this identity as a choice of implementation. Under these conditions, the following updates can be used for mini-batch OMF-B algorithm. Update for $x_{v_t}$ reads as,
\begin{align}\label{updateMBX}
x_{v_t} = (C_t^\top C_t)^{-1} C_t^\top y_{v_t}
\end{align}
and update rule for $C_t$ can be given as,
\begin{align}\label{updateMBC}
C_t &= (\lambda C_{t-1} + y_{v_t} x_{v_t}^\top) (\lambda I + x_{v_t} x_{v_t}^\top)^{-1}
\end{align}
which is no longer same as the Broyden's rule for mini-batch observations.
\subsection{Handling Missing Data}
\begin{algorithm}[t]
\begin{algorithmic}[1]
\caption{OMF-B with Missing Data}\label{MissingOMFB}
\State Initialise $C_0$ randomly and set $t = 1$.
\Repeat
\State Pick $k_t \in [n]$ at random.
\State Read $y_{k_t} \in \bR^m$
\For{$\text{Iter} = 1:2$}
\begin{align*}
x_{k_t} =& \left((M_{C_t} \odot C_t)^\top (M_{C_t} \odot C_t)\right)^{-1} \times \\
&(M_{C_t} \odot C_t)^\top (m_{k_t}\odot y_{k_t}) \\
C_t &= C_{t-1} + \frac{(m_{k_t} \odot (y_{k_t} - C_{t-1}x_{k_t})) x_{k_t}^\top}{\lambda + x_{k_t}^\top x_{k_t}}
\end{align*}
\EndFor
\State $t \leftarrow t+1$
\Until{convergence}
\end{algorithmic}
\end{algorithm}
In this subsection, we give the update rules which can handle the missing data. We only give the updates for single data vector observations because deriving the mini-batch update for missing data is not obvious and also become computationally demanding as $|v_t|$ increases. So we only consider the case $|v_t| = 1$ i.e. we assume observing only a single-column at a time.

We define a mask $M \in \{0,1\}^{m\times n}$, and we denote the data matrix with missing entries with $M \odot Y$ where $\odot$ denotes the Hadamard product. We need another mask to update related entries of the estimate of the dictionary matrix $C_t$, which is denoted as $M_{C_t}$ and naturally, $M_{C_t} \in \{0,1\}^{m\times r}$. Suppose we have an observation $y_{k_t}$ at time $t$ and some entries of the observation are missing. We denote the mask vector for this observation as $m_{k_t}$ which is $k_t$'th column of $M$. We construct $M_{C_t}$ for each $t$ in the following way:
\begin{align*}
M_{C_t} = \underbrace{[m_{k_t},\ldots, m_{k_t}]}_{r \textnormal{ times}}.
\end{align*}
The use of $M_{C_t}$ stems from the following fact. We would like to solve the following least squares problem for $x_{k_t}$ (for fixed $C_t$),
\begin{align}\label{LSmissing}
\min_{x_{k_t}} \big\| m_{k_t} \odot \left(y_{k_t} - C_t x_{k_t}\right) \big\|_2^2.
\end{align}
One can easily verify that,
\begin{align*}
m_{k_t} \odot \left(C_t x_{k_t}\right) = \left(M_{C_t} \odot C_t \right) x_{k_t}.
\end{align*}
Then \eqref{LSmissing} can equivalently be written as,
\begin{align*}
\min_{x_{k_t}} \big\| \left(m_{k_t} \odot y_{k_t}\right) - \left( M_{C_t} \odot C_t\right) x_{k_t} \big\|_2^2.
\end{align*}
As a result the update rule for $x_{k_t}$ becomes the following pseudoinverse operation,
\begin{align*}
x_{k_t} =& ((M_{C_t} \odot C_t)^\top (M_{C_t} \odot C_t))^{-1} \times \\
&(M_{C_t} \odot C_t)^\top (m_{k_t}\odot y_{k_t}),
\end{align*}
and the update rule for $C_t$ (for fixed $x_{k_t}$) can trivially be given as,
\begin{align*}
C_t &= C_{t-1} + \frac{(m_{k_t} \odot (y_{k_t} - C_{t-1}x_{k_t})) x_{k_t}^\top}{\lambda + x_{k_t}^\top x_{k_t}}.
\end{align*}
We denote the results on dataset with missing entries in Experiment~\ref{ExperimentMissing}.
\section{Experimental Results}\label{SecExperiment}
In this section, we demonstrate two experiments on the Olivetti faces dataset\footnote{Available at: \url{http://www.cs.nyu.edu/~roweis/data.html}} consists of $400$ faces with size of $64\times 64$ grayscale pixels. We first compare our algorithm with stochastic gradient descent matrix factorization in the sense of error vs. runtimes. In the second experiment, we randomly throw away the \%25 of each face in the dataset, and try to fill-in the missing data. We also compare our results with NMF \cite{LeeSeungNMF}.
\subsection{Comparison with stochastic gradient MF}
In this section, we compare our algorithm with the stochastic gradient descent matrix factorization (SGMF) algorithm \cite{sismanisSGD}. Notice that one can write the classical matrix factorization cost as,
\begin{align*}
\big\| Y - W H\big\|_F^2 = \sum_{k=1}^n \big\| y_k - W h_k \big\|_2^2
\end{align*}
so it is possible to apply alternating stochastic gradient algorithm \cite{sismanisSGD}. We derive and implement the following updates for SGMF,
\begin{align*}
W_t &= W_{t-1} - \gamma^W_t \nabla_{W} \big\| y_{k_t} - W h_{k_t}\big\|_2^2 \Big|_{W = W_{t-1}} \\
h^{t}_{k_t} &= h^{t-1}_{k_{t}} - \gamma^{h}_t \nabla_{h} \big\| y_k - W_t h \big\|_2^2 \Big|_{h = h^{t-1}_{k_{t}}}
\end{align*}
for uniformly random $k_t \in [n]$ for each $t$. The following conditions hold for convergence: $\sum_{t=1}^\infty \gamma^W_t = \infty$ and $\sum_{t=1}^\infty \left(\gamma^W_t\right)^2 < \infty$ and same conditions hold for $\gamma_t^h$. In practice we scale the step-sizes like $\alpha/t^{\beta}$ where $0<\alpha<\infty$ and $0.5 < \beta < 1$. These are other parameters we have to choose for both $W$ and $h$. It is straightforward to extend this algorithm to mini-batches \cite{sismanisSGD}. We merely replace $k_t$ with $v_t$.
\begin{figure}[t]
\begin{center}
\includegraphics[scale=0.5]{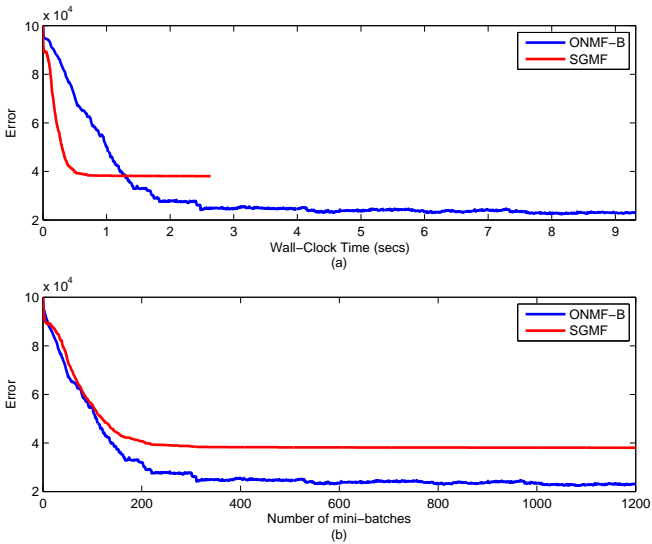}
\end{center}
\caption{Comparison with SGMF on Olivetti faces dataset. (a) This plot shows that although SGMF is faster than our algorithm (since we employ two iterations for each mini-batch), and SGMF processes the dataset in a much less wall-clock time, we achieve a lower error in the same wall-clock time. (b) This plot shows that our algorithm uses samples in a more efficient manner. We obtain lower errors for the same processed amount of data.}
\label{SGMF}
\end{figure}

In this experiment, we set identical conditions for both algorithms, where we use the Olivetti faces dataset, set $r = 30$, and use mini-batch size $10$ for both algorithms. We have carefully tuned and investigated step-size of the SGMF to obtain the best performance. We used scalar step sizes for the matrix $W$ and we set a step-size for each mini-batch-index, i.e. we use a matrix step-size for updating $h_{v_t}$. We set $\lambda = 10$. At the end, both algorithms passed $30$ times over the whole dataset taking mini-batch samples at each time. We measure the error by taking Frobenius norm of the difference between real data and the approximation.

The results are given in Fig.~\ref{SGMF}. We compared error vs. runtimes and observed that SGMF is faster than our algorithm in the sense that it completes all passes much faster than OMF-B as can be seen from Fig.~\ref{OMFB}(a). However our algorithm uses data much more efficiently and achieves much lower error rate \textit{at the same runtime} by using much fewer data points than SGMF. In the long run, our algorithm achieves a lower error rate within a reasonable runtime. Additionally, our algorithm has a single parameter to tune to obtain different error rates. In contrast, we had to carefully tune the SGMF step-sizes and even decay rates of step-sizes. Compared to SGMF, our algorithm is much easier to implement and use in applications.
\subsection{Handling missing data on Olivetti dataset}\label{ExperimentMissing} 
\begin{figure}[t]
\begin{center}
\includegraphics[scale=0.18]{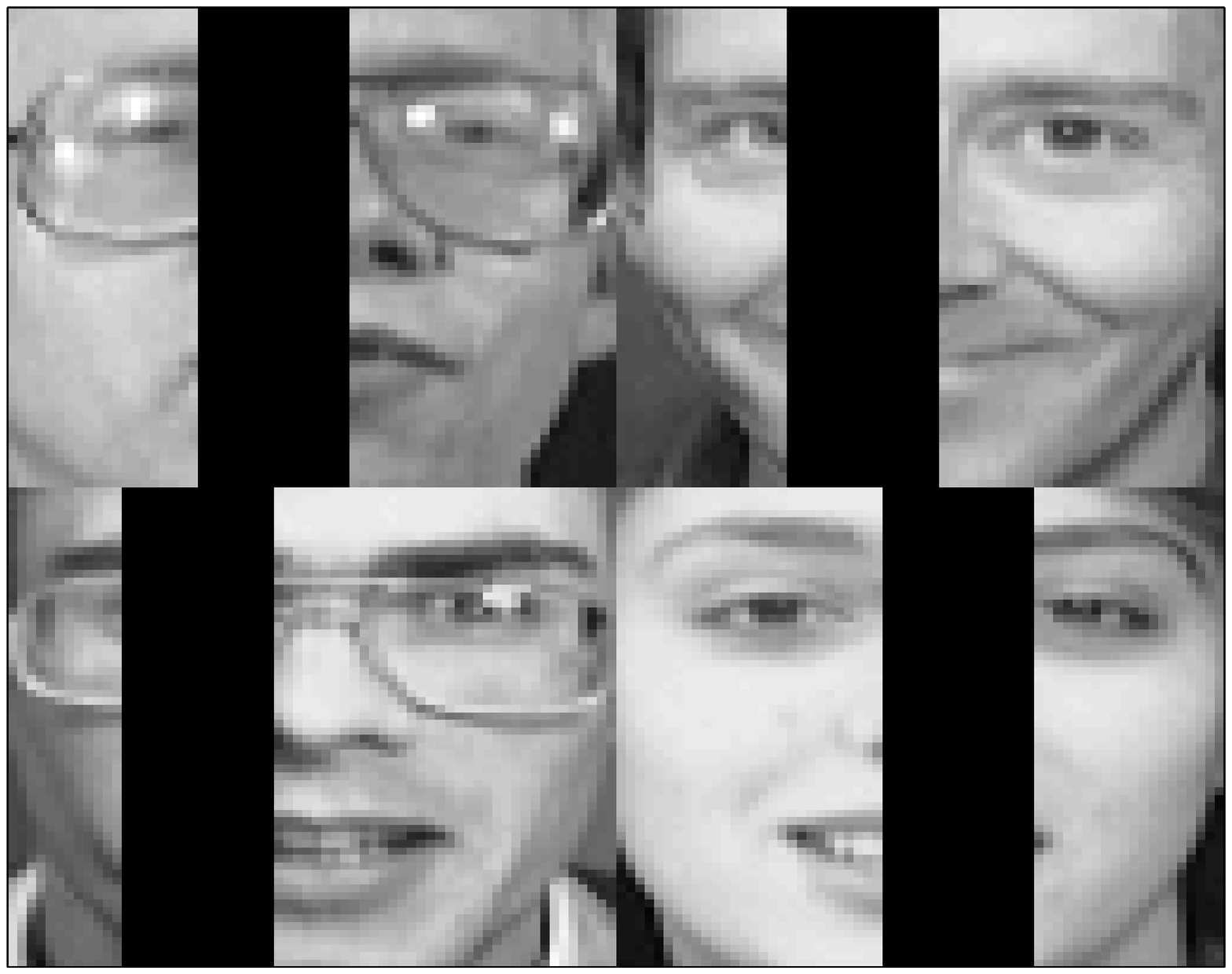}
\includegraphics[scale=0.18]{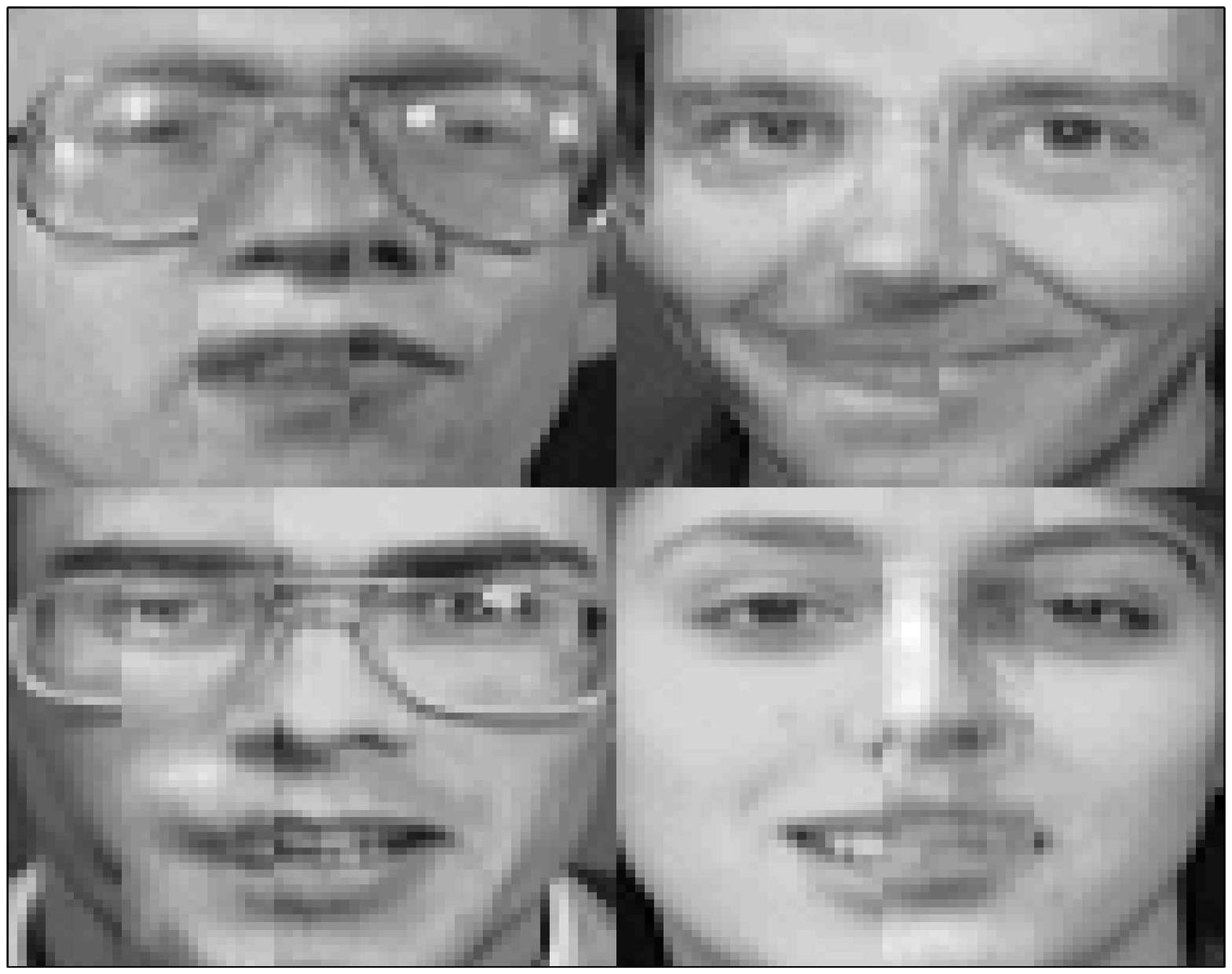}
\includegraphics[scale=0.18]{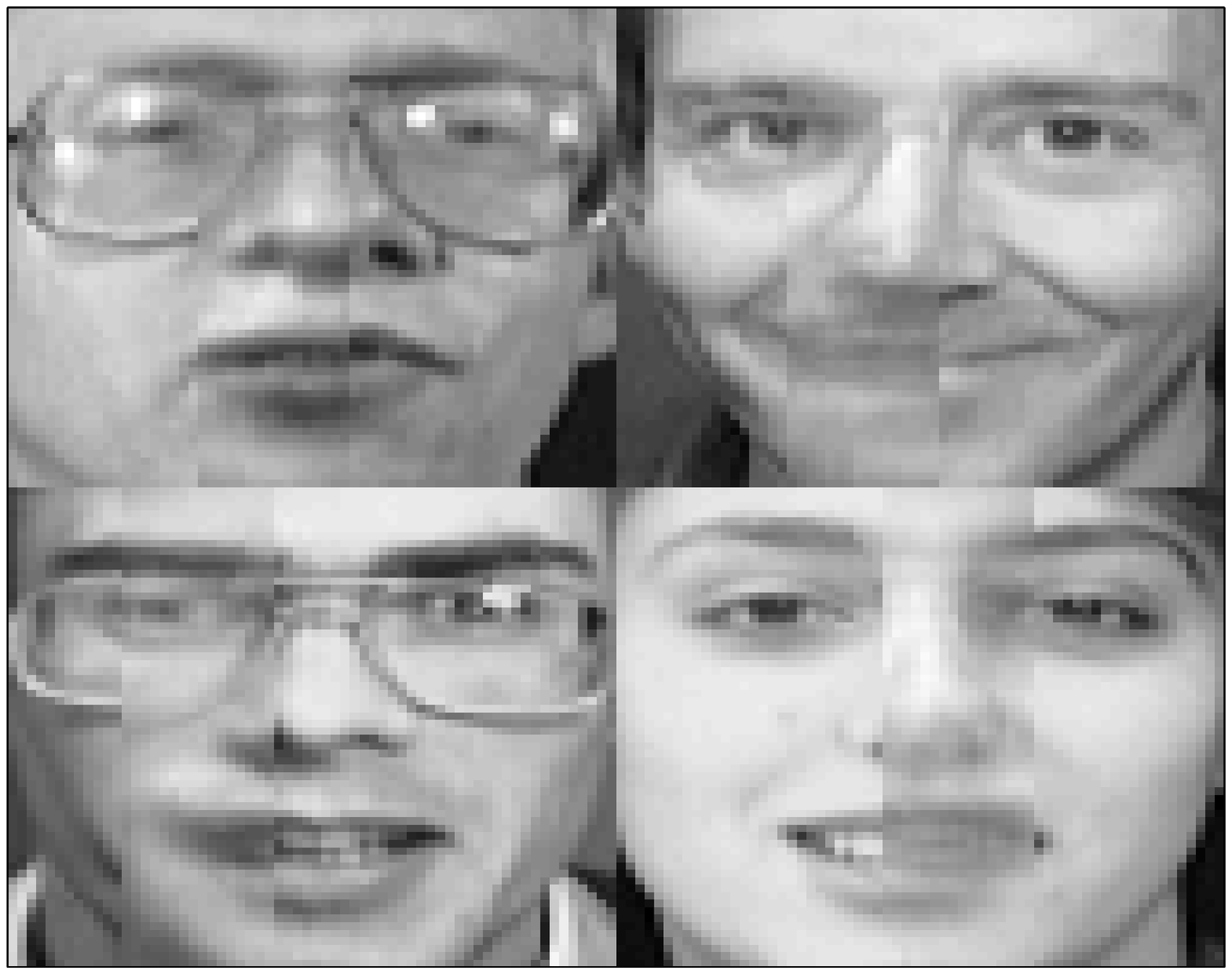}
\end{center}
\caption{A demonstration on Olivetti faces dataset consists of 400 faces of size $64\times 64$ with \%25 missing data. Some example faces with missing data are on the left. Comparison of results of OMF-B (middle) with 30 online passes over dataset and NMF with 1000 batch iterations (right). Signal-to-noise ratios (SNR) are: OMF-B: 11.57, NMF: 12.13 where initial SNR is 0.75.}
\label{figFaces}
\end{figure}

In this experiment, we show results on the Olivetti faces dataset with missing values where \%25 of the dataset is missing (we randomly throw away \%25 of the faces). Although this dataset is small enough to use a standard batch matrix factorisation technique such as NMF, we demonstrate that our algorithm competes with NMF in the sense of Signal-to-Noise Ratio (SNR). We compare our algorithm with NMF in terms of number of passes over data vs. SNR. We choose $\lambda = 2$, and set inner iterations as $2$. Our algorithm achieves approximately same SNR values with NMF (1000 batch passes over data) with only 30 online passes over dataset. This shows that our algorithm needs much less low-cost passes over dataset to obtain comparable results with NMF. Numbers and visual results are given in Fig.~\ref{figFaces}.
\section{Conclusions and Future Work}\label{SecConc}
We proposed an online and easy-to-implement algorithm to compute matrix factorizations, and demonstrated results on the Olivetti faces dataset. We showed that our algorithm competes with the state-of-the-art algorithms in different contexts. Although we demonstrated our algorithm in a general setup by taking random subsets of the data, it can be used in a sequential manner as well, and it is well suited to streaming data applications. In the future work, we plan to develop probabilistic extensions of our algorithm using recent probabilistic interpretations of quasi-Newton algorithms, see e.g. \cite{hennig2013quasi} and \cite{hennig2015probabilistic}. The powerful aspect of our algorithm is that it can also be used with many different priors on columns of $X$ such as the one proposed in \cite{piecewiseNMF}. As a future work, we think to elaborate more complicated problem formulations for different applications.
\section*{Acknowledgements}
The author is grateful to Philipp Hennig for very helpful discussions. He is also thankful to Taylan Cemgil and S. Ilker Birbil for discussions, and to Burcu Tepekule for her careful proofreading. This work is supported by the TUBITAK under the grant number 113M492 (PAVERA).
\section*{Appendix}
We derive $\nabla_{C_t} G_{k_t}$ as the following. First we will find $\nabla_{C_t} \big\|y_{k_t} - C_t x_{k_t}\big\|_2^2$ which is the derivative of the first term. Notice that
\begin{align*}
\big\|y_{k_t} - C_t x_{k_t} \big\|_2^2 = \Tr\left(y_{k_t}^\top y_{k_t} - 2 y_{k_t}^\top C_t x_{k_t} + x_{k_t}^\top C_t^\top C_t x_{k_t}\right)
\end{align*}
First of all the first term is not important for us, since it does not include $C_t$. Using standard formulas for derivatives of traces \cite{matrixcookbook}, we arrive,
\begin{align}\label{derivativeOfFirst}
\nabla_{C_t} \big\|y_{k_t}-C_t x_{k_t} \big\|_2^2 = -2 y_{k_t} x_{k_t}^\top + 2 C_t x_{k_t} x_{k_t}^\top
\end{align}
The second term of the cost function can be written as,
\begin{align*}
\lambda \big\|C_t - C_{t-1} \big\|_F^2 = \lambda \Tr\left((C_t - C_{t-1})^\top (C_t - C_{t-1})\right)
\end{align*}
If we take the derivative with respect to $C_t$ using properties of traces \cite{matrixcookbook},
\begin{align}\label{derivativeOfSecond}
\nabla_{C_t} \lambda\big\|C_t - C_{t-1} \big\|_F^2 = 2\lambda C_t - 2\lambda C_{t-1}
\end{align}
By summing \eqref{derivativeOfFirst} and \eqref{derivativeOfSecond}, setting them equal to zero, and leaving $C_t$ alone, one can show \eqref{beforeShermanMorrison} easily. Using Sherman-Morrison formula, one can obtain the update rule given in the Eq.~\eqref{updateC}.
\ifCLASSOPTIONcaptionsoff
  \newpage
\fi


%
\bibliographystyle{IEEEtran}
\bibliography{draft}

\end{document}